\documentclass{article}

\usepackage[preprint]{neurips_2020}

\usepackage[utf8]{inputenc} 
\usepackage[T1]{fontenc}    
\usepackage{hyperref}       
\usepackage{url}            
\usepackage{booktabs}       
\usepackage{amsfonts}       
\usepackage{nicefrac}       
\usepackage{microtype}      

\usepackage{chngcntr}
\usepackage{amsmath}
\usepackage{amsthm}
\usepackage{caption}
\usepackage{wrapfig}
\usepackage{fancyhdr,graphicx,amsmath,amssymb}
\usepackage[ruled,vlined]{algorithm2e}
\include{pythonlisting}

\usepackage{xcolor}

\newcommand{\eps}{\epsilon}
\renewcommand{\l}{\lambda}

\renewcommand{\part}{\partial}
\newcommand{\f}{\frac}
\newcommand{\N}[1]{\left\lVert #1 \right\rVert}

\renewcommand{\cal}{\mathcal}

\DeclareMathOperator*{\argmax}{arg\,max}

\newcommand{\m}{\begin{bmatrix}}
\newcommand{\mq}{\end{bmatrix}}

\newtheorem{definition}{Definition}[section]

\newtheorem{theorem}{Theorem}
\newtheorem{lemma}[theorem]{Lemma}

\title{Adaptive Online Planning \\ for Continual Lifelong Learning}

\author{
  Kevin Lu \\
  UC Berkeley \\
  \texttt{kzl@berkeley.edu} \\
  \And
  Igor Mordatch \\
  Google Brain \\
  \texttt{imordatch@google.com} \\
  \And
  Pieter Abbeel \\
  UC Berkeley \\
  \texttt{pabbeel@cs.berkeley.edu} \\
}

\begin{document}

\maketitle

\begin{abstract}
We study learning control in an online reset-free lifelong learning scenario, where mistakes can compound catastrophically into the future and the underlying dynamics of the environment may change. Traditional model-free policy learning methods have achieved successes in difficult tasks due to their broad flexibility, but struggle in this setting, as they can activate failure modes early in their lifetimes which are difficult to recover from and face performance degradation as dynamics change. On the other hand, model-based planning methods learn and adapt quickly, but require prohibitive levels of computational resources. We present a new algorithm, Adaptive Online Planning (AOP), that achieves strong performance in this setting by combining model-based planning with model-free learning. By approximating the uncertainty of the model-free components and the planner performance, AOP is able to call upon more extensive planning only when necessary, leading to reduced computation times, while still gracefully adapting behaviors in the face of unpredictable changes in the world -- even when traditional RL fails.
\end{abstract}

\section{Introduction}

We consider agents in a lifelike setting, where agents must simultaneously act and learn in the world continuously with limited computational resources. All decisions are made online; there are no discrete episodes. Furthermore, the world is vast -- too large to feasibly explore exhaustively -- and changes over the course of the agent's lifetime, like how a robot's actuators might deteriorate with continued use. Mistakes are costly, as they compound downstream; there are no resets to wipe away past errors. To perform well at reasonable computational costs, the agent must utilize its past experience alongside new information about the world to make careful, yet performant, decisions.

Non-stationary worlds require algorithms that are fundamentally robust to changes in dynamics. Factors that would lead to a change in the environment may either be too difficult or principally undesirable to model: for example, humans might interact with the robot in unpredictable ways, or furniture in a robot's environment could be rearranged. Therefore, we assume that the world can change unpredictably in ways that cannot be learned, and focus on developing algorithms that instead handle these changes gracefully, without using extensive computation.

Model-based planning is useful for quickly learning control, but is computationally demanding and can be biased by the finite planning horizon. Model-free RL is sample inefficient, but capable of cheaply accessing past experience. Consequently, we would like to distill expensive experience generated by a powerful planner into model-free networks to reduce computation. However, deciding on when to use planning versus model-free control is nontrivial. When uncertain about a course of action, humans use an elongated model-based search to evaluate long-term trajectories, but fall back on habitual behaviors learned with model-free paradigms when uncertainty is low \cite{evans1984, daw2005uncertainty, dayan2014model, kahn2003}. By measuring this uncertainty, we can make informed decisions about when to use extensive planning.

\begin{figure*}[h]
    \centering
    \includegraphics[width=1\linewidth]{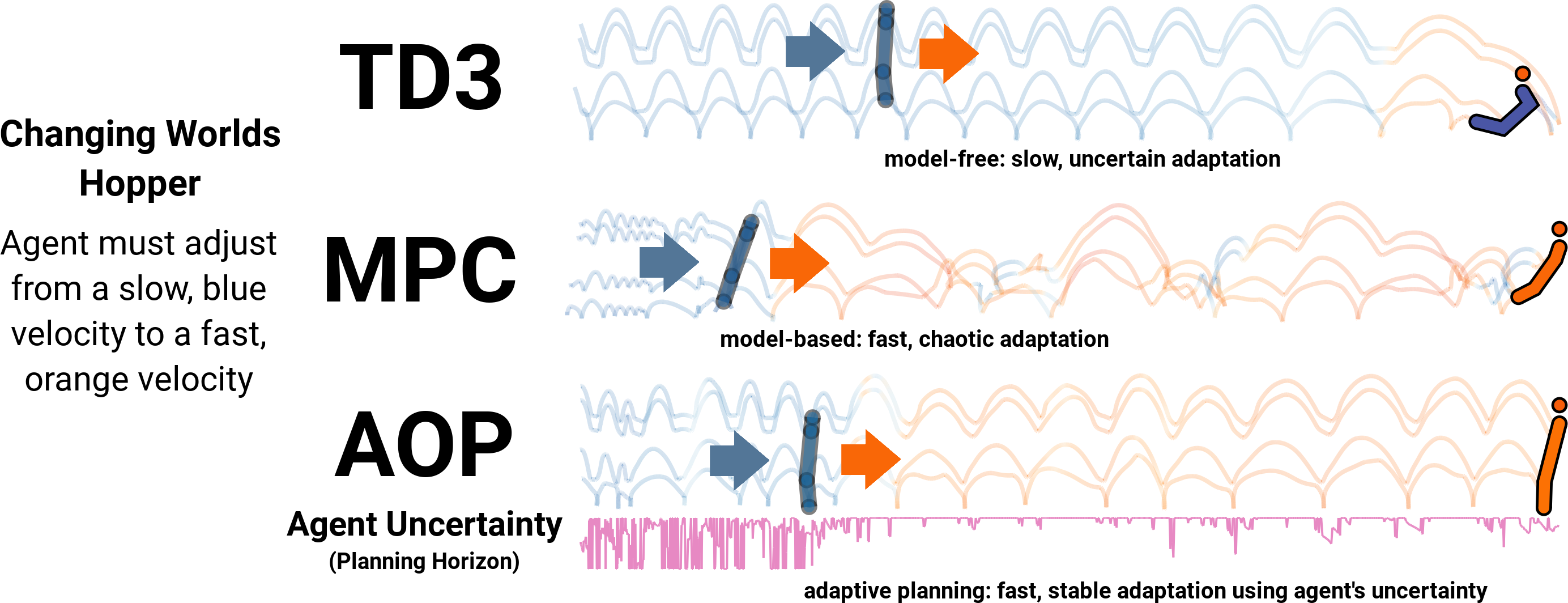}
    \caption{Traces of Hopper positions. An unobserved increase in the target velocity is encountered at the times marked by the blue shadows. TD3 adapts slowly to the target speed and catastrophically falls over. MPC adapts quickly, but chaotically. AOP is able to both effectively and rapidly adapt. For more information, see our website: \url{https://sites.google.com/berkeley.edu/aop}.}
    \label{fig:aopcart}
\end{figure*}

Our approach combines model-based planning with model-free learning, along with an adaptive computation mechanism, to tackle this setting. Like a robot that is well-calibrated when first coming out of a factory, we give the agent access to a ground truth dynamics model that lacks information about future changes to the dynamics, such as different settings in which the robot may be deployed. This allows us to make progress on finite computation continual learning without complications caused by model learning. The dynamics model is updated immediately at world changes. However, as we show empirically, knowing the dynamics alone falls far short of success at this task.

We present a new algorithm, Adaptive Online Planning (AOP), that links model-based planning with model-free policy optimization. We interpolate between the two methods using a unified update rule formulation that is amenable to reduced computation when combined with a switching mechanism. We inform this mechanism with the uncertainty given by an ensemble of value functions. We show that access to the ground truth model is not sufficient by itself, as PPO \cite{ppo} and TD3 \cite{td3} perform poorly, even with such a model. In particular, we discover that policy-based methods degrade in performance over time as the world changes. We provide intuition, qualitative analysis, and demonstrate empirically that AOP is capable of integrating the two methodologies to reduce computation while achieving and maintaining strong performance in non-stationary worlds, outperforming other model-based planning methods and avoiding the performance degradation of policy learning methods.

\section{Background}

\subsection{Notation and Setting}

We consider the world as an infinite-horizon MDP defined by the tuple $\cal{M} = \{\cal{S}, \cal{A}, \cal{R}, \cal{T}, \gamma\}$, where $\cal{S}$ is the state space, $\cal{A}$ is the action space, $\cal{R}: \cal{S} \times \cal{A} \to [-R_{\text{max}}, R_{\text{max}}]$ is the reward function, $\cal{T}: \cal{S} \times \cal{A} \times \cal{S} \to \mathbb{R}$ are the transition probabilities, and $\gamma \in [0,1)$ is the discount factor. The world changes over time: $(\cal{T}, \cal{R})$ may switch to new $(\cal{T}', \cal{R}')$ at multiple points during training. Unlike traditional RL, the agent's state is not reset at these world changes, and these changes are not observed in the state. The agent has access to a perfect local model, i.e. it can generate rollouts using the current $(\cal{T}, \cal{R})$ starting from its current state $s_t \in \cal{S}$. It does not have access to future $(\cal{T}', \cal{R}')$. The agent's goal is to execute a policy $\pi(a_t | s_t)$, that maximizes the $T$-horizon expected return as $J^\pi_T(s_t) = \mathbb{E}_{\tau\sim\pi}[\sum_{k=0}^{T-1} \gamma^k r(s_t, a_t)]$ for $T = \infty$. We denote the value function $V^\pi(s) = J^\pi_\infty(s)$.

We consider reset-free online learning to be the setting where both acting and learning must occur on a per-timestep basis, and there are no episodes that reset the state. At each timestep, the agent must execute its training, and is then forced to immediately output an action. The world changes between environments sequentially. Agents should use past experience to perform well in new tasks, while avoiding past negative transfer of experience. In addition to the these adversities, agents must avoid failure sink states that prevent future learning progress, as the agent cannot reset. These challenges yield a difficult setting which more closely parallels a human lifetime than traditional episodic RL.

\subsection{Model-Based Planning} \label{sec:mpc}

Online model-based planning (see Algorithm \ref{alg:mpc}) evaluates future sequences of actions using a model of the environment, develops a projected future trajectory over some time horizon, and then executes the first action of that trajectory, before replanning at the next timestep. We specifically focus on shooting-based Model Predictive Control (MPC). When the update rule is a softmax weighting, this procedure is called Model Predictive Path Integral (MPPI) control \cite{mppi}. Due to the nature of this iterative and extended update, this procedure is computationally expensive, and can lead to unpredictable and irregular behaviors, as it is a brute-force black box optimizer.

\subsection{Model-Free Policy Optimization}

Model-free algorithms encode the agent's past experiences in neural network functions dependent only on the current state, often in the form of a value function critic and/or policy actor. As a result, such algorithms can have difficulty learning long-term dependencies and struggle early on in training; temporally-extended exploration is difficult. In exchange, they attain high asymptotic performance, having shown successes in a variety of tasks in the episodic setting \cite{dqn, ppo}. As a consequence of their compact nature, once learned, these algorithms tend to generate cyclic and regular behaviors, whereas model-based planners have no such guarantees (see Figure \ref{fig:aopcart} for an illustration).

We run online versions of TD3 \cite{td3} and PPO \cite{ppo} as baselines to AOP. While there is no natural way to give a policy access to the ground truth model, we allow the policies to train on future trajectories generated via the ground truth model, similarly to model-based policy optimization algorithms \cite{steve, metrpo}, in order to help facilitate fair comparisons to model-based planners (see Algorithm \ref{alg:td3}).

\section{Unifying Model-Based Planning and Model-Free Control}

\subsection{Update Rule Perspective on Model-Based Planning vs Policy Optimization} \label{sec:ppg}

From a high-level perspective, the model-based planning and model-free policy optimization procedures are very similar (see Algorithms \ref{alg:mpc} and \ref{alg:td3} for a side-by-side comparison). Where the planner generates noisy rollouts to synthesize a new trajectory, the model-free algorithm applies noise to the policy to generate data for learning. After an update step, either an action from the planned trajectory or one call of the policy is executed. These procedures are only distinct in their respective update rules: planning uses a fast update, while policy optimization uses a slow one.

\noindent\begin{minipage}[t][][b]{\textwidth}
    \centering
    \begin{minipage}{.48\textwidth}
        \centering
        \IncMargin{1em}
        \begin{algorithm}[H] \label{alg:mpc}
            \SetAlgoLined
            Initialize action trajectory $\tau_{plan}$ \\
            \While{alive} {
                Generate $n$ rollouts based on $\tau_{plan}$ \\
                Use rollouts to update $\tau_{plan}$ \\
                Execute first action of $\tau_{plan}$
            }
        \caption{Model-Based Planning}
        \end{algorithm}
    \end{minipage}
    \begin{minipage}{.48\textwidth}
        \centering
        \IncMargin{1em}
        \begin{algorithm}[H] \label{alg:td3}
            \SetAlgoLined
            Initialize policy $\pi_\phi$ \\
            \While{alive} {
                Generate $n$ rollouts based on $\pi_\phi$ \\
                Use rollouts to update $\pi_\phi$ \\
                Execute an action from $\pi_\phi$
            }
        \caption{Policy Optimization}
        \end{algorithm}
    \end{minipage}
\end{minipage}

The primary contribution of our algorithm, AOP, is unifying both update rules to compensate for their individual weaknesses. AOP distills the learned experience from the planner into the off-policy learning method of TD3 and a value function, so that planning and acting can be cheaper in the future.

\subsection{Model-Based Planning with Terminal Value Approximation} \label{sec:polo}

AOP uses MPPI \cite{mppi} as the planning algorithm, with a learned terminal value function $\hat{V}^\pi(s)$, seeking to optimize the $H$-step objective $\hat{J}^\pi_\infty(\tau^{(H)}) = \sum_{k=0}^{H-1} \gamma^k r(s_k, a_k) + \gamma^H \hat{V}^\pi(s_H)$. This process is repeated for several iterations, and then the first action is executed in the environment. $\hat{V}$ is generated by an ensemble of $n$ value functions as $\hat{V}(s) = \f{1}{\kappa} \log \frac{1}{N} \sum_{i=1}^n e^{\kappa V_i(s)}$, similar to POLO \cite{polo}. The value ensemble improves the exploration of the optimization procedure \cite{rpf, bs-dqn}. The log-sum-exp function serves as a softmax, enabling optimistic exploration. The $\frac{1}{n}$ term normalizes the estimate to lie between the mean and the max of the ensemble, determined by the temperature hyperparameter $\kappa$, ensuring that the approximation is semantically meaningful for value estimation.

Crucially, instead of a fixed horizon, AOP adaptively chooses the planning horizon to interpolate between model-free control (short $H$) and model-based planning (long $H$); see Algorithm \ref{alg:aop}. This planning horizon is set based on the uncertainty of the value function, discussed in Section \ref{sec:thor}.

\begin{wrapfigure}{r}{0.4\linewidth}
    \begin{algorithm}[H]
        \caption{AOP} \label{alg:aop}
        \While{alive} {
            Generate $\tau_\pi$ from prior $\pi_\theta$ \\
            $\tau_{plan} = \argmax_{\tau_\pi, \tau_{plan}} \hat{J}^\pi_\infty(\tau)$ \\
            Select time horizon $H_t$ \\
            \For{$k\gets1$ \KwTo $max\_iters$} {
                Update $\tau_{plan}$ with MPC \\
                \If{$\Delta < \Delta_{thres}$} {
                    Stop planning with probability $1 - \eps_{plan}$
                }
            }
            Execute first action of $\tau_{plan}$
        }
    \end{algorithm}
\end{wrapfigure}

\subsection{Off-Policy Model-Free Prior} \label{sec:priors}

We use TD3 \cite{td3} as a prior $\pi_\theta$ to the planning procedure, with the policy learning off of the data generated by the planner, which allows the agent to recall past experience quickly. Note that we use $\pi_\theta$ to distinguish from $\pi$, which represents the agent as a whole. In line with past work \cite{rajeswaran2017learning, zhu2018reinforcement}, we found that imitation learning can cap the asymptotic performance of the learned policy. As a baseline, we also run behavior cloning (BC) as a prior, and refer to the resulting algorithms as AOP-TD3 and AOP-BC.

We note that MPC and policy optimization are both special cases of AOP. MPC is equivalent to AOP with a constant setting for the time horizon that always uses full planning iterations ($\Delta_{thres} = 0$). Policy optimization is equivalent to AOP with one planning iteration, since the first plan is a noisy version of the policy, acting as the data collection procedure in standard policy learning.

\section{Adaptive Planning for Reset-Free Lifelong Learning}

\subsection{Long-Term Regret in Lifelong Learning} \label{sec:ltr}


In our setting, we can consider ``long-term regret'' derived from planning over the model using a limited time horizon. A more formal/detailed version is included in Appendix \ref{sec:moreltr}.

\begin{definition}
    \label{def:regret}
    Let the planning regret denote the suboptimality in the infinite-horizon value of the rollout when using an MPC planner over an $H$-step horizon and executing the policy afterwards vs. planning using an infinite-horizon planner: $R(s) = J^*_\infty(s) - (J^\pi_H(s) + \gamma^H V^\pi(s_H))$.
\end{definition}

Using a $^*$ to denote states generated by an optimal policy/planner, we can rewrite the regret as:

{\centering
$ \displaystyle
\begin{aligned}
    R(s) &= \sum_{i=0}^\infty \gamma^i r(s^*_i, \pi^*(s^*_i)) - (\sum_{i=0}^{H-1} \gamma^i r(s_i, \pi(s_i)) + V^\pi(s_H)) \\
        &= \gamma^H (V^*(s^*_H) - V^\pi(s_H)) + \sum_{t=0}^{H-1} \gamma^t (r(s^*_t, a^*_t) - r(s_t, a_t)) \\
        &= \gamma^H LR(s) + SR(s)
\end{aligned}
$
\par
}

where in the last line we decompose the regret as the sum of the \emph{short-term} and \emph{long-term} regret. We focus on the long-term regret term due to how it models our lifelong learning setting.

\theoremstyle{plain}
\begin{lemma}
    \label{lemma:longtermregret}
    Let the value approximation error be bounded as $\max_s \left| \hat{V}^\pi(s) - V^\pi(s) \right| \leq \epsilon_V$ and the future agent policy suboptimality be bounded as $\max_s \left| V^*(s) - V^\pi(s) \right| \leq \epsilon_P$. Then we have:
    
    {\centering
    $ \displaystyle
        LR(s) \leq \frac{2 R_{\text{max}}}{\gamma^H (1 - \gamma)} + 2 \epsilon_V + \epsilon_P
    $
    \par
    }
    
\end{lemma}

At a high level, this intuitively poses the long-term regret as the sum of short-term greed, long-term value approximation error, and long-term agent policy suboptimality.

\subsubsection{Why Focus on Long-Term Regret?}

The notion of long-term regret is particularly important in the reset-free setting, as there exists ``pseudo-terminal'' sink states where, not only does the agent incur large penalty, but also it is hard to make further learning progress at all. Traditional episodic RL only incurs the large penalty, and does not suffer a block on the learning ability, hence the traditional notion of ``long-term'' refers to only within the current trajectory, but the reset-free notion of ``long-term'' refers to the entire training loop, which makes improper handling of long-term regret far more catastrophic. This also corresponds to how we consider $H$ to be far shorter than the total lifelong horizon, or similarly $\gamma$ or $\gamma^H \approx 1$.

Consider the trajectory of TD3 shown in Figure \ref{fig:aopcart}. One-step lookahead policy optimization algorithms suffer a high value approximation error at dramatic world changes. At the end of the shown trajectory, the policy optimizes for the short-term ability to achieve the new orange target velocity, in exchange for falling over, which inhibits its ability to continue learning and acting, thereafter suffering a large long-term penalty.  While this is not easily modeled by standard notions of regret or the short-term regret, this is captured by the long-term regret, and is a principle cause of policy degradation. To a lesser extent, the MPC trajectory also exhibits this, as it continually greedily optimizes for reward in the near future, matching the target velocity extremely well, but suffering periods of failure where it falls over and does not match the target velocity, though it generally recovers sufficiently.



\subsection{Adaptive Planning Horizon} \label{sec:thor}

Now, under Lemma \ref{lemma:longtermregret}, we can consider AOP as a tradeoff between these two extremes. In particular, we propose AOP as minimizing $H$ given a constraint on the value approximation error/long-term regret, allowing the agent to utilize more model-free control and save computation as long as the value function approximator is accurate. We use two heuristics to approximate the uncertainty:

\begin{itemize}
    \item The standard deviation of the value function ensemble, which represents the epistemic uncertainty of the value of the state \cite{rpf}. In particular, we automatically use the full planning horizon if the standard deviation is greater than a threshold $\sigma_{thres}$.
    
\begin{equation} \label{eq:belerr}
    \eps(H | \tau_k) = (\hat{J}^{\pi}_{\infty}(\tau^{(H_{full} - H)}_k) - \f{1}{n} \sum_{i=1}^n V_i(s_{t+H}))^2
\end{equation}
    
    \item The Bellman error of the value (see Equation \ref{eq:belerr}), which directly approximates the error of the value function. In particular, this uncertainty is high in changing worlds, whereas the standard deviation is not. We use the longest $H \leq H_{full}$ such that $\eps(H | \tau_k) > \eps_{thres}$.
\end{itemize}

While choices for $\sigma_{thres}$ and $\eps_{thres}$ are somewhat arbitrary, we show in Appendix \ref{app:aophyperparameters} that AOP is not particularly sensitive to them, and they generally transfer between environments.

\subsection{Early Planning Termination} \label{sec:ept}

Additionally, we save computation by running less planning iterations. Past model-based planning procedures \cite{pets,poplin} run a fixed number of iterations of MPC per timestep before executing an action in the environment, which can be wasteful, as later iterations are less useful. Consequently, we propose to decide on the number of planning iterations on a per-timestep basis. After generating a new trajectory $\tau_{k+1}$ from the $k$-th iteration of planning, we measure the improvement $\Delta(\tau_{k+1} | \tau_k) = \f{\hat{R}(\tau_{k+1}) - \hat{R}(\tau_k)}{|\hat{R}(\tau_k)|}$ against the trajectory $\tau_k$ of the previous iteration. When this improvement decreases below a threshold $\Delta_{thres}$, we terminate planning for the current timestep with probability $1-\eps_{plan}$. We find that using a stochastic termination rule allows for more robustness against local minima.

\section{Empirical Evaluations}

We investigate several questions empirically:

\begin{enumerate}
    \item What are the challenges in the continual lifelong learning setting? When developing further algorithms in this setting, what should we focus on?
    
    \item How does AOP perform in well-known states, novel states, and in changing worlds scenarios? How do traditional on-policy and off-policy methods fare in these situations?
    
    \item Are the variance and the Bellman error of the value ensemble suitable metrics for representing the agent's uncertainty and determining the planning computational budget?
\end{enumerate}

\subsection{Lifelong Learning Environments}

We propose four environments to evaluate our proposed algorithm in the continual lifelong learning setting: Hopper, Humanoid, Ant, and Maze. We consider three classes of experiments: ``standard'' experiments (S) where the world never changes, which is similar to standard RL except for not having resets; novel states (NS) experiments where the dynamics $\cal{T}$ never changes, but the agent's assigned task changes and is observed (equating to new states); and changing worlds experiments (CW) -- the focus of this work -- where the dynamics $\cal{T}$ and/or task will change and not be observed in the state.

\textbf{Hopper:} A hopping agent is rewarded based on how closely its forward velocity matches an unobserved changing target velocity. Long-term regret is extremely poignant in this environment, as the hopping motion can cause long-term value to be sacrificed for short-term reward, and momentum can cause the agent to fall over, after which it can be hard to recover, stifling further learning progress. Hopper is not a difficult control environment, but it is the most unstable environment in this setting. In the novel states setting, the target velocity is included in the observation.

\textbf{Humanoid:} A humanoid agent seeks to achieve a fixed forward velocity, where the mass of its right arm varies greatly during its lifetime. Recovering after falling face-first is extremely difficult.

\textbf{Ant}: A four-legged agent seeks to achieve a fixed forward velocity, where a joint at random is disabled every 2000 timesteps. Again, falling over can hinder learning greatly.

\textbf{Maze:} We test in a 2D continuous point mass maze, where the agent seeks to reach a goal, which is included in the observation. We consider two versions: (1) a novel states Maze, where the walls of the maze remain constant, but new goals are introduced periodically, and (2) a changing worlds Maze, where both the walls and the goal constantly change. We also test both versions in a both a dense reward and a sparse reward setting.

These environments test a variety of challenges a lifelong agent might face: changes in the reward specification (Hopper, Maze), changes in the underlying dynamics parameters (Humanoid), changes in the action space (Ant), and long-term global changes in the world structure (Maze). Furthermore, they each exhibit difficult local optima that can be difficult to recover from in the reset-free setting.

\subsection{Baselines and Ablations}

We run AOP-BC, POLO, MPC, TD3, and PPO as baselines against AOP-TD3; they can be seen as ablations/special cases of our proposed algorithm.  We consider two versions of MPC, with $8$ and $3$ planning iterations (referred to as MPC-8 and MPC-3), to consider the effect of computation.

\begin{figure*}[h]
    \centering
    \includegraphics[scale=1.4]{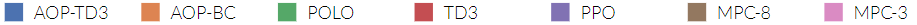}
    \begin{minipage}{.24\textwidth}
        \centering
        \includegraphics[scale=0.22]{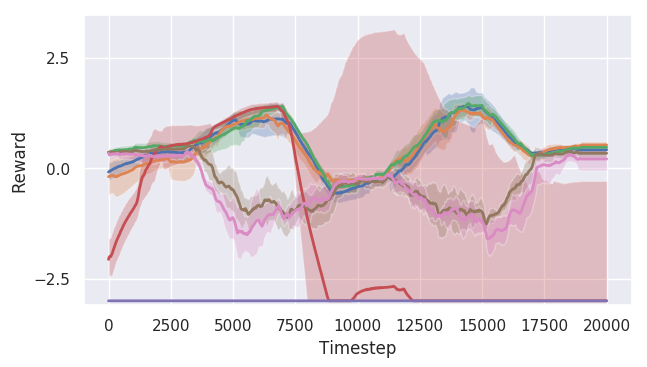} \\
        (a) Hopper
    \end{minipage}
    \begin{minipage}{.24\textwidth}
        \centering
        \includegraphics[scale=0.22]{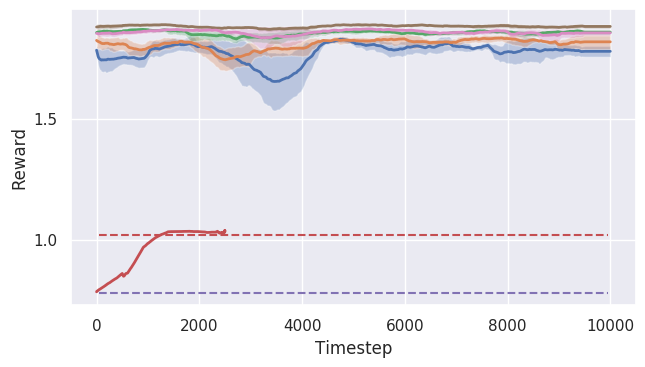} \\
        (b) Humanoid
    \end{minipage}
    \begin{minipage}{.24\textwidth}
        \centering
        \includegraphics[scale=0.22]{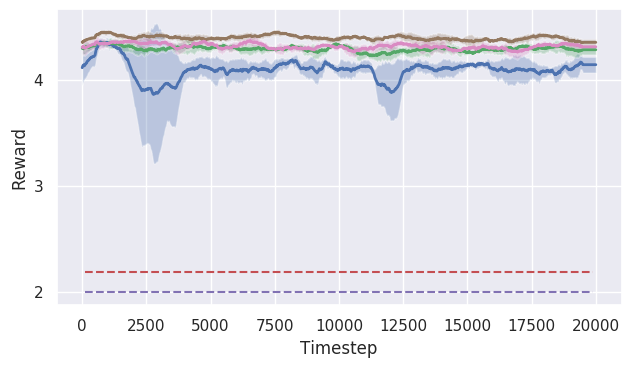} \\
        (c) Ant
    \end{minipage}
    \begin{minipage}{.24\textwidth}
        \centering 
        \includegraphics[scale=0.22]{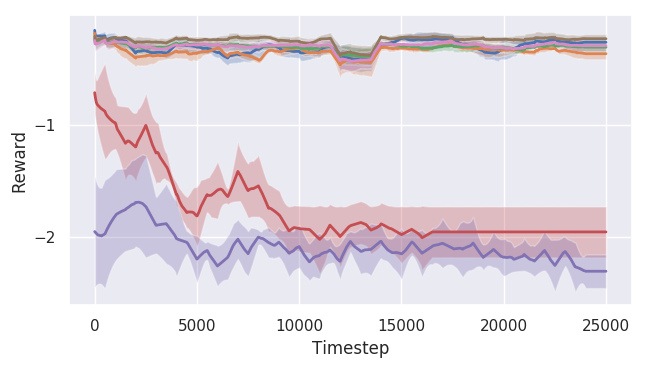} \\
        (d) Maze (D)
    \end{minipage}
    
    \caption{Reward curves for changing worlds environments. Rewards are for a single timestep, not an episode. Note that some worlds may be more difficult than others, and yield a naturally lower reward. The results are averaged over 5 seeds; the shaded area depicts one standard deviation. AOP achieves competitive performance with strong model-based planners. Policy-only control methods degrade in performance as worlds change.}
    \label{fig:crew}
\end{figure*}

\begin{table*}[h]
    \caption{Average lifetime rewards. (D) and (S) denote dense/sparse reward. We show the average for 5 seeds with two standard deviations. Strongest results are bolded, determined by increases in reward that correspond to semantically meaningful asymptotic differences. We didn't fully run the seeds marked with a $^*$ because they fail early, but they typically converge to close to the shown values.}
    \label{table:rew}
    \begin{center}
    \begin{tabular}{ccccccccc}
    \multicolumn{1}{c}{} & \multicolumn{1}{c}{\bf Env} & \multicolumn{1}{c}{\bf AOP-TD3} & \multicolumn{1}{c}{\bf AOP-BC} & \multicolumn{1}{c}{\bf POLO} & \multicolumn{1}{c}{\bf TD3} & \multicolumn{1}{c}{\bf PPO} & \multicolumn{1}{c}{\bf MPC-8} & \multicolumn{1}{c}{\bf MPC-3}
    \\ \hline \\
    S & Hopper    & \bf{0.12 $\pm$ 0.16} & \bf{0.33 $\pm$ 0.22} & \bf{0.51} & \bf{0.23} & -14.4 & \bf{0.36} & \bf{0.19} \\
    NS & Hopper   & \bf{0.41 $\pm$ 0.18} & \bf{0.53 $\pm$ 0.18} & \bf{0.59} & \bf{0.40} & -14.2 & -0.28 & -0.49 \\
    CW & Hopper   & \bf{0.48 $\pm$ 0.24} & \bf{0.45 $\pm$ 0.12} & \bf{0.57} & -2.42 & -13.1 & -0.30 & -0.48 \\
    S & Human     & \bf{1.79 $\pm$ 0.02} & \bf{1.80 $\pm$ 0.02} & \bf{1.86} & 1.01$^*$ & 0.78$^*$ & \bf{1.88} & \bf{1.87} \\
    CW & Human    & \bf{1.78 $\pm$ 0.03} & \bf{1.80 $\pm$ 0.02} & \bf{1.85} & 1.01$^*$ & 0.78$^*$ & \bf{1.88} & \bf{1.86} \\
    S & Ant       & \bf{4.15 $\pm$ 0.11} & \bf{4.17 $\pm$ 0.03} & \bf{4.30} & 2.19 & n/a & \bf{4.41} & \bf{4.34} \\
    CW & Ant      & \bf{4.10 $\pm$ 0.06} & \bf{4.15 $\pm$ 0.03} & \bf{4.29} & 2.05 & n/a & \bf{4.40} & \bf{4.32} \\
    NS & Maze (D) & \bf{-0.21 $\pm$ 0.08} & \bf{-0.25 $\pm$ 0.02} & \bf{-0.25} & -1.81 & -2.14 & \bf{-0.19} & \bf{-0.25} \\
    CW & Maze (D) & \bf{-0.29 $\pm$ 0.07} & \bf{-0.34 $\pm$ 0.03} & \bf{-0.30} & -1.17 & -2.10 & \bf{-0.19} & \bf{-0.30} \\
    NS & Maze (S) & \bf{0.85 $\pm$ 0.07} & 0.70 $\pm$ 0.06 & 0.62 & -0.68 & -0.88 & 0.69 & 0.61 \\
    CW & Maze (S) & \bf{0.69 $\pm$ 0.20} & 0.56 $\pm$ 0.04 & 0.57 & -0.66 & -0.74 & 0.58 & 0.52 \\
    \end{tabular}
    \end{center}
\end{table*}

\begin{table*}[h]
    \caption{Timesteps rolled out by planner (planning levels) as a fraction of MPC-8 for MPC-based algorithms. Shown are the average and the range (min-max) across the environments.}
    \label{table:plan}
    \begin{center}
    \begin{tabular}{llllll}
    \multicolumn{1}{c}{\bf AOP-TD3} & \multicolumn{1}{c}{\bf AOP-BC} & \multicolumn{1}{c}{\bf POLO} & \multicolumn{1}{c}{\bf MPC-8} & \multicolumn{1}{c}{\bf MPC-3}
    \\ \hline \\
    \bf{11.99\%} (1.40\% - 16.62\%) & \bf{11.96\%} (2.86\% - 15.17\%) & 37.50\% & 100\% & 37.50\%
    \end{tabular}
    \end{center}
\end{table*}

\subsection{Challenges in Reset-Free Continual Lifelong Learning}
\label{sec:challenges}

Rewards are shown in Figure \ref{fig:crew} and Table \ref{table:rew}, and planning is in Table \ref{table:plan}. AOP plans only at $1-17\%$ of MPC-8, but performs comparably or better, successfully learning the desired behavior on all tasks.

\textbf{Reset-Free Setting:} Even with model access, these environments are challenging. In the standard episodic reinforcement learning setting, long-term action dependencies are learned from past experience over time, and this experience can be utilized when the agent resets to the initial state. However, in the online reset-free setting, these dependencies must be learned on the fly, and if the agent falls, it must return to the prior state in order to use that information, which is often harder than the original task; the model-free algorithms generally fail to overcome this challenge of long-term regret.

\begin{figure}[h]
    \centering
    \includegraphics[scale=1.4]{pics/algo_labels.png}
    \includegraphics[scale=0.4]{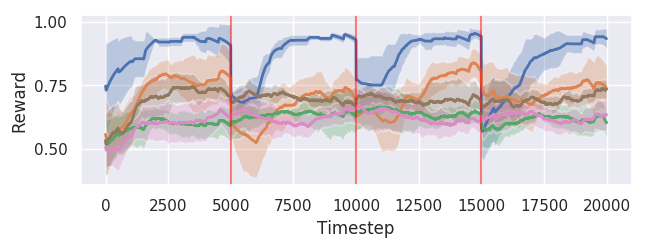}
    \includegraphics[scale=0.4]{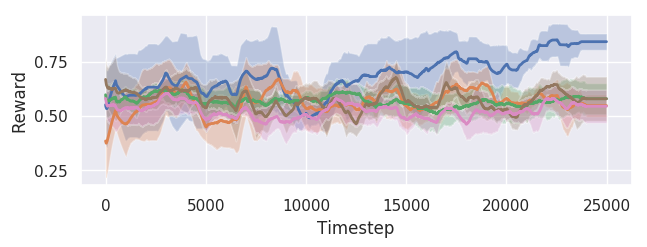} 
    \caption{Sparse maze rewards. Left: novel states (red lines are new tasks). Right: changing worlds.}
    \label{fig:sparse}
\end{figure}

\textbf{Vast Worlds:} In the sparse mazes, MPC is significantly outperformed by AOP-TD3 (Figure \ref{fig:sparse}), and the model-free algorithms struggle to make any progress at all, showing their lackluster exploration. Even POLO -- the exploration mechanism of AOP -- faces weaker performance, indicating that AOP-TD3 has not only correctly identified when planning is important, but is able to effectively leverage additional computation to increase its performance whilst still using less overall computation.

The additional performance in the sparse novel states Maze over MPC-8 also shows AOP's ability to consolidate experience to improve performance in situations it has seen before. Furthermore, in the sparse changing worlds Maze, the performance of AOP improves over time, indicating that AOP has learned value and policy functions for effective forward transfer in a difficult exploration setting.

\textbf{Policy Degradation:} TD3's performance significantly degrades in the changing worlds settings, as does PPO's (see Figure \ref{fig:crew}). PPO, an on-policy method, struggles in general. In the novel states Hopper, where the policy is capable of directly seeing the target velocity, TD3 performs very well, even learning to outperform MPC. However, without the help of the observation, in the changing worlds, TD3's performance quickly suffers after world changes, incurring large long-term regret. The model-based planning methods naturally do not suffer this degradation; AOP is able to maintain its performance and computational savings, even through many world changes, despite its reliance on model-free components. This phenomenon is detailed more in Appendix \ref{sec:polbehavior}.

\subsection{Analysis of AOP Planning and Uncertainty Approximation}

\begin{figure}[h]
    \centering
    \includegraphics[scale=0.27]{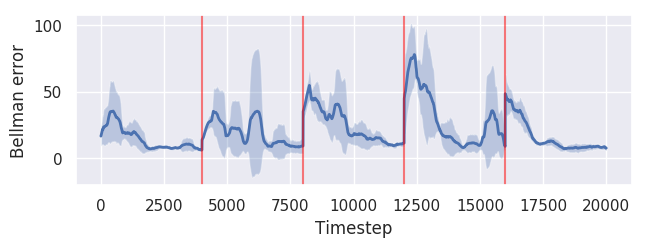} \includegraphics[scale=0.27]{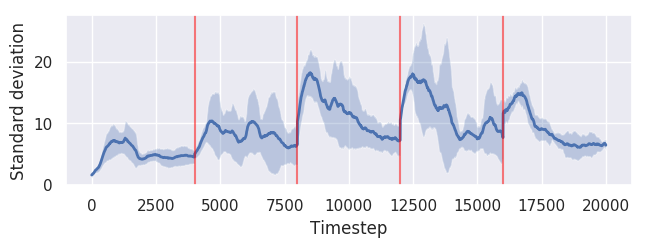}
    \includegraphics[scale=0.27]{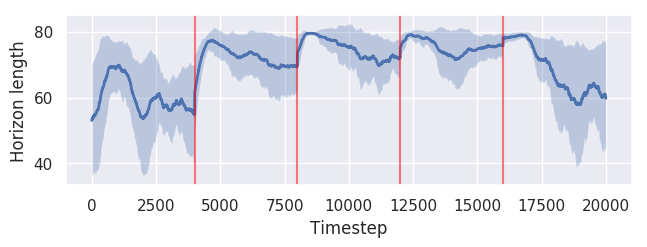}
    \caption{Value uncertainty for AOP in changing worlds Hopper. Red lines denote world changes. Uncertainty decreases as the agent learns to act in each world.}
    \label{fig:hopnc}
\end{figure}

The standard deviation and Bellman error over time of AOP for the changing worlds Hopper is shown in Figure \ref{fig:hopnc}. After each world change, the Bellman error spikes, and then decreases as time goes on and the agent becomes more familiar with the environment. These trends are reflected in the time horizon, which decreases as the agent trains in each world, indicating that the standard deviation and Bellman error are suitable metrics for considering agent uncertainty.

\begin{figure}[h]
    \centering
    \includegraphics[scale=0.2]{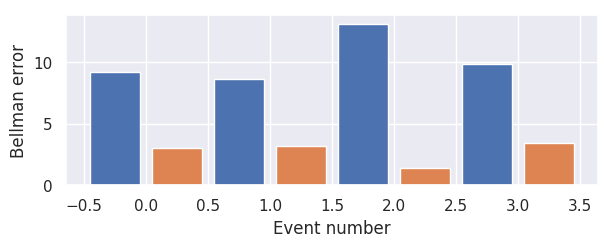} \includegraphics[scale=0.2]{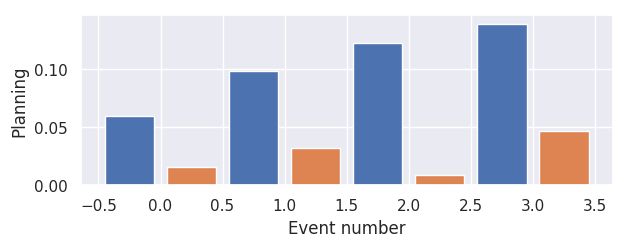}
    \includegraphics[scale=0.2]{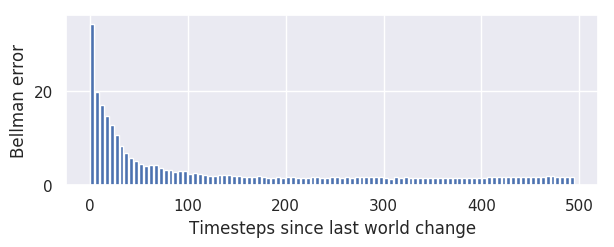} \includegraphics[scale=0.2]{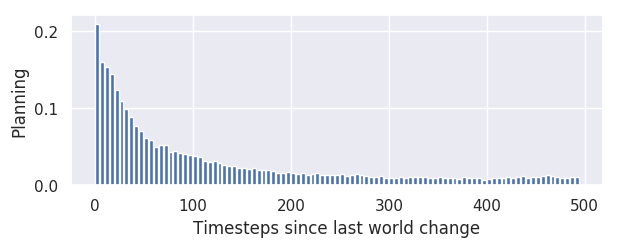}
    \caption{Maze (D). From left to right: (1; NS) average Bellman error of the first occurrence of a task (blue) vs the last (orange), (2; NS) average number of planning steps, (3; CW) average Bellman error by the time since the last world change, and (4; CW) average number of planning steps. The amount of planning significantly decreases in later visits/timesteps.}
    \label{fig:maze_ns}
\end{figure}

Figure \ref{fig:maze_ns} shows AOP behavior in the dense Mazes. When encountering novel states, Bellman error is high, but as time progresses, when confronted with the same states again, the Bellman error is low. The number of planning timesteps matches this: AOP correctly identifies a need to plan early on, but greatly saves computation later, when it is immediately able to know the correct action with almost no planning. The same effect occurs when measuring the time since the world changed for the changing worlds. At the beginning of a new world, the amount of planning is high, before quickly declining to nearly zero, almost running with the speed of a policy: $\approx 100 \times$ faster than MPC-8.

\section{Related Work}

\textbf{Continual learning:} Most past continual learning work \cite{finn2019onlinemeta, goodfellow2013forget, kirkpatrick17forgetting, parisi, rolnick18replay, ruvolo13ella, schwarz18compress} has focused on catastrophic forgetting, which AOP is resilient to (see Appendix \ref{sec:polbehavior}), but was not a primary focus of our work, because unlike traditional continual learning literature, the same state/input can denote different dynamics and tasks, since they are not embedded in the observation, which makes the notion of catastrophic forgetting poorly defined. In contrast, our setting emphasizes the desire to quickly perform any new task well with limited regret, which naturally includes previous tasks, but does not focus on them. In the supervised framework of \cite{vandeven19scenarios}, our changing worlds setting is classified as Domain-IL, with our novel states variant being classified as the easier Task-IL. \cite{mole} learns multiple models to represent different tasks for continual learning in episodic RL. In contrast to previous notions of regret \cite{alquier16regret, besbes15stoopt, yang16dynregret}, we consider a notion of regret that focuses on how well the agent can learn in the future as a consequence of near short-term actions.

\textbf{Planning and learning:} Algorithms that combine planning with learning have been studied with great variety in both discrete and continuous domains \cite{exit, pets, polo, poplin}. Our algorithm is closest to POPLIN \cite{poplin}, which uses MPC with a learned policy prior. \cite{levine2013guided, mordatch2015interactive} propose to use priors that make the planner stay close to policy outputs, which is problematic in changing worlds, when the policy is not accurate. \cite{sezener20computation} proposes to use learned upper confidence bounds to evaluate the value of computation for the discrete MCTS algorithm, which is a similar notion to our work. \cite{mctsnets} generalizes the MCTS algorithm and proposes to learn the algorithm instead; having the algorithm set computation levels could be interesting in our setting. \cite{azziz2018, pets, mbmf, poplin} learn models of the environment to perform MPC. \cite{mbmpo, modtrust, metrpo} utilize model ensembles to reduce model overfitting for policy learning.

\section{Conclusion}

We proposed AOP, incorporating model-based planning with model-free learning, and introduced environments for evaluating algorithms in a new continual lifelong learning setting. We highlighted the shortcomings of traditional RL, analyzed the performance of and signals from AOP's model-free components, and showed experimentally that AOP was able to reduce computation while achieving high performance in difficult tasks, competitive with a significantly more powerful MPC procedure.

\section{Broader Impact}

Our work focuses on how we can design agents that are capable of learning in a setting where the training and evaluation environments are not distinct. Most modern applications and work in reinforcement learning (and machine learning systems more broadly) are constrained by a notion of training in a specific setting or dataset and then being applied in a real-world evaluation setting. This can be problematic in that robotic systems can encounter novel situations where the model is uncertain of how to act; this is a well-known bottleneck of deep learning applications and, for example, can be catastrophic in safety-critical scenarios, such as self-driving cars. Agents should learn to adapt gracefully to such scenarios, training while acting. Our work is a step in the direction of a setting where we have biologically-inspired, intelligent agents that can continuously learn in their lifetime in the same manner that a human learns, adapting to novel and changing situations with minimal long-term risk. However, there is still work to be done, in this setting and in reinforcement learning more broadly, to understand control in safety-critical scenarios. A potential negative impact of this work could result from abuses of RL use cases, where algorithms may be applied to situations where the decision making capability is not well understood or constrained, which could lead to user overconfidence and misuse. More intelligent agents should have strong notions of causality and hierarchy, which is currently a fundamental limitation of deep learning systems.

\subsubsection*{Acknowledgments}

We would like to thank Vikash Kumar, Aravind Rajeswaran, Michael Janner, and Marvin Zhang for helpful feedback, comments, and discussions.

\bibliographystyle{plain}
\bibliography{sources}

\newpage

\title{test}
\maketitle

\begin{center}
    \textbf{\Large Appendix}
\end{center}

\appendix
\counterwithin{figure}{section}
\counterwithin{table}{section}

\section{Further Discussion of Long-Term Regret} \label{sec:moreltr}

In this section, we provide more details, commentary, and the proof for the analysis from Section \ref{sec:ltr}.

\subsection{Definitions}

Recall that we define the regret as the suboptimality of the MPC planner when using an $H$-step horizon against an optimal infinite-horizon MPC policy, where actions for timesteps after time $H$ are given by the agent policy $\pi$, representing the agent's ability to replan later and decide on those actions. Note that this $\pi$ is not the learned parameterized policy, but representing the agent as a whole.

Specifically, when the dynamics are deterministic (simplifying the equations), as in this work, the MPC policy is optimized as the following objective over the ground-truth model:

\begin{equation}
    \pi(a_t | s_t) = \argmax_{\pi} \hat{J}^\pi_\infty(s_t) = \argmax_{\pi} \sum_{k=0}^{H-1} \gamma^k r(s_{t+k}, \pi(s_{t+k})) + \gamma^H \hat{V}^\pi(s_{t+H})
\end{equation}

In pure MPC, $\hat{V}^\pi(s) = 0$, which is problematic when $H$ is much smaller than the lifetime horizon. For our algorithms, we use a learned terminal value function parameterized as a neural network. In shooting-based MPC, as used in this work, the MPC policy simply takes the form of open-loop actions, although the result holds for more general forms of closed-loop optimization.

Our formulation of regret (from Definition \ref{def:regret}) yields the following decomposition:

\begin{equation}
    R(s_t) = \gamma^H LR(s_t) + SR(s_t)
\end{equation}

where the long-term and short-term regret are defined as:

\begin{equation}
    LR(s_t) = V^*(s^*_{t+H}) - V^\pi(s_{t+H})
\end{equation}

\begin{equation}
    SR(s_t) = \sum_{k=0}^{H-1} \gamma^k (r(s^*_{t+k}, a^*_{t+k}) - r(s_{t+k}, a_{t+k}))
\end{equation}

Note that the short-term regret is not necessarily positive, and likely is negative from a practical perspective, as MPC will prefer to greedily optimize for return in the first $H$ steps, particularly when $\hat{V}^\pi(s)$ inaccurately models the true value of the future states.

\subsection{Effect of Planning Horizon on Long-Term Regret}

\begin{figure}[h]
    \centering
    \includegraphics[scale=0.5]{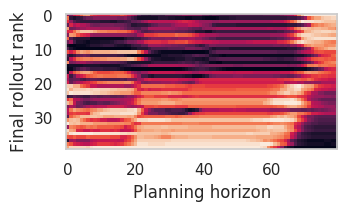}
    \caption{Pitfalls of using a finite-horizon planner. The pure MPC planner is only able to correctly rank the trajectories over a long planning horizon. The data shown here was from a planning iteration of MPC-3 on the Hopper environment.}
    \label{fig:mpcpitfall}
\end{figure}

In Figure \ref{fig:mpcpitfall}, we show the evaluation of each trajectory in the sampled population of an MPC iteration. Each row represents a trajectory, and are ordered vertically in terms of the strongest trajectory at the top, and the weakest at the bottom, where we run a more extended simulation to approximate the total return for the trajectories. Each column represents a timestep of planning, e.g. the left of the figure shows short-horizon planning and the right shows long-horizon planning. The pixel at cell (row $i$, column $j$) represents the ranking of trajectory $i$ when evaluated by MPC with a horizon of $j$; white denotes the strongest rankings, and purple/black denote the weakest rankings. An ideal MPC with an accurate value approximator is able to order the trajectories perfectly, which is approximately done when planning with the long horizon ($H = 80$) on the far right.

Figure \ref{fig:mpcpitfall} indicates the issue with only short-term planning, and a key reason why reset-free environments (particularly in environments with momentum or short-term rewards like Hopper) are difficult. In the presence of a weak (or absent) value function approximator, incorrect assignment of values to trajectories can help lead to catastrophic mistakes.

\subsection{Proof of Lemma \ref{lemma:longtermregret}}

Let $\tau^* = (s^*_t, a^*_t, s^*_{t+1}, a^*_{t+1}, ..., s^*_{t+H})$ denote the optimal trajectory generated by $\infty$-horizon planning and $\tau = (s_t, a_t, s_{t+1}, a_{t+1}, ..., s_{t+H})$ denote the trajectory generated by $H$-horizon planning.

Because the optimum of the $H$-horizon planner is realized with $\tau$, the planner evaluation incorrectly assigns $\hat{J}^*_\infty(s_t) \leq \hat{J}^\pi_\infty(s_t)$. Substituting for the individual terms, this yields:

\begin{align*}
    \sum_{k=0}^{H-1} \gamma^k r(s^*_{t+k}, \pi(s^*_{t+k})) + \gamma^H \hat{V}^\pi(s^*_{t+H})
        \leq \sum_{k=0}^{H-1} \gamma^k r(s_{t+k}, \pi(s_{t+k})) + \gamma^H \hat{V}^\pi(s_{t+H})
\end{align*}

Rearranging and bounding as the maximum difference in the sum of rewards:

\begin{align*}
     \hat{V}^\pi(s^*_{t+H})- \hat{V}^\pi(s_{t+H}) &\leq \frac{1}{\gamma^H} \sum_{k=0}^{H-1} \gamma^k (r(s^*_{t+k}, \pi(s^*_{t+k})) - r(s_{t+k}, \pi(s_{t+k}))) \\
        &\leq \frac{1}{\gamma^H} \sum_{k=0}^{H-1} \gamma^k (R_{\text{max}} - (-R_{\text{max}})) \\
        &= \frac{2 R_{\text{max}} (1 - \gamma^H)}{\gamma^H (1 - \gamma)}
\end{align*}

Now, we can use this result and substitute the assumed bounds to derive the result:

\begin{align*}
    LR(s_t) &= V^*(s^*_{t+H}) - V^\pi(s_{t+H}) \\
        &= (V^*(s^*_{t+H}) - V^\pi(s^*_{t+H})) + V^\pi(s^*_{t+H}) - V^\pi(s_{t+H}) \\
        &\leq \eps_P + V^\pi(s^*_{t+H}) - V^\pi(s_{t+H}) \\
        &\leq \eps_P + (\hat{V}^\pi(s^*_{t+H}) + \eps_V) - (\hat{V}^\pi(s_{t+H}) - \eps_V) \\
        &\leq \frac{2 R_{\text{max}} (1 - \gamma^H)}{\gamma^H (1 - \gamma)} + 2 \eps_V + \eps_P
\end{align*}

\subsection{Remarks on Lemma \ref{lemma:longtermregret}}

We note the following characteristics of the above three inequalities:

\begin{enumerate}
    \item The first inequality, bounding long-term policy suboptimality, follows from the agent being unable to execute an optimally, and is greatest when the agent is unable to progress through learning at all. More powerful procedures (like MPC) are less subject to this, and this is amplified when ``pseudo-terminal'' states are present in the MDP.
    
    \item The second inequality, bounding the value function approximator, follows from overestimation of the desired state and underestimation of the optimal state.
    
    \item The third inequality, bounding the short-term greedy optimization, is tightest when the MDP allows for high short-term gain in exchange for longer-term penalty, like the Hopper environment, and the MPC planning procedure optimally solves for the maximum, and is weaker when the planning procedure is softer (ex. using a procedure like MPPI), or starts with a long-term prior (like a policy) and then performs regularized local optimization.
\end{enumerate}

\section{Episodic Policy Performance and Policy Degradation} \label{sec:polbehavior}

\begin{figure}[h]
    \centering
    \includegraphics[scale=0.4]{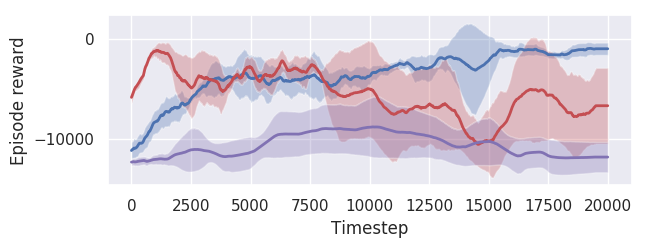}
    \includegraphics[scale=0.4]{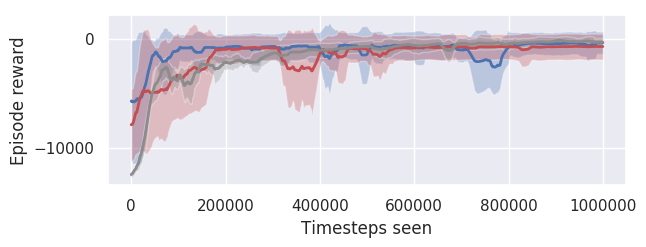}
    
    \caption{Left: Policy episode performance for changing worlds Hopper. Shown is the performance of policy only (no planner) throughout training. Right: Performance on initial starting state and target velocity after additional training with TD3 (blue: AOP-TD3, red: TD3, gray: from scratch).}
    \label{fig:pol_performance}
\end{figure}

Here we expand on our consideration of policy degradation from Section \ref{sec:challenges}. In Figure \ref{fig:pol_performance}, we plot an episodic reward of the policy running from the initial starting state after each timestep (for the current target velocity), which can be considered to be the standard reinforcement learning setting with a different training regimen. Note that since the AOP policy is learned from off-policy data (the planner), it suffers from divergence issues and should be weaker than TD3 on its own \cite{fujimoto2018divergence}.

Matching the result in Figure \ref{fig:crew} (a), the TD3 (red) policy degrades in performance over time, but the AOP (blue) policy does not, despite TD3 learning much faster initially. Because the optimization procedure is the same, this result suggests that the policy degradation effect likely stems from exploration, rather than from an issue with optimization or the network falling into a parameter ``trap''.

Figure \ref{fig:pol_performance} shows the result of tuning the policy learned by AOP after seeing every target velocity once (blue) vs. by TD3 (red) vs. training a new policy from scratch (gray), learning via the standard episodic TD3 algorithm on the first target velocity. The AOP policy learns faster, showing that it is capable of quick backward transfer and adaptation, while TD3 by itself is comparable to starting over.

\section{Additional Algorithm Specification} \label{app:algs}

\begin{figure}[h]
    \centering
    \includegraphics[scale=0.5]{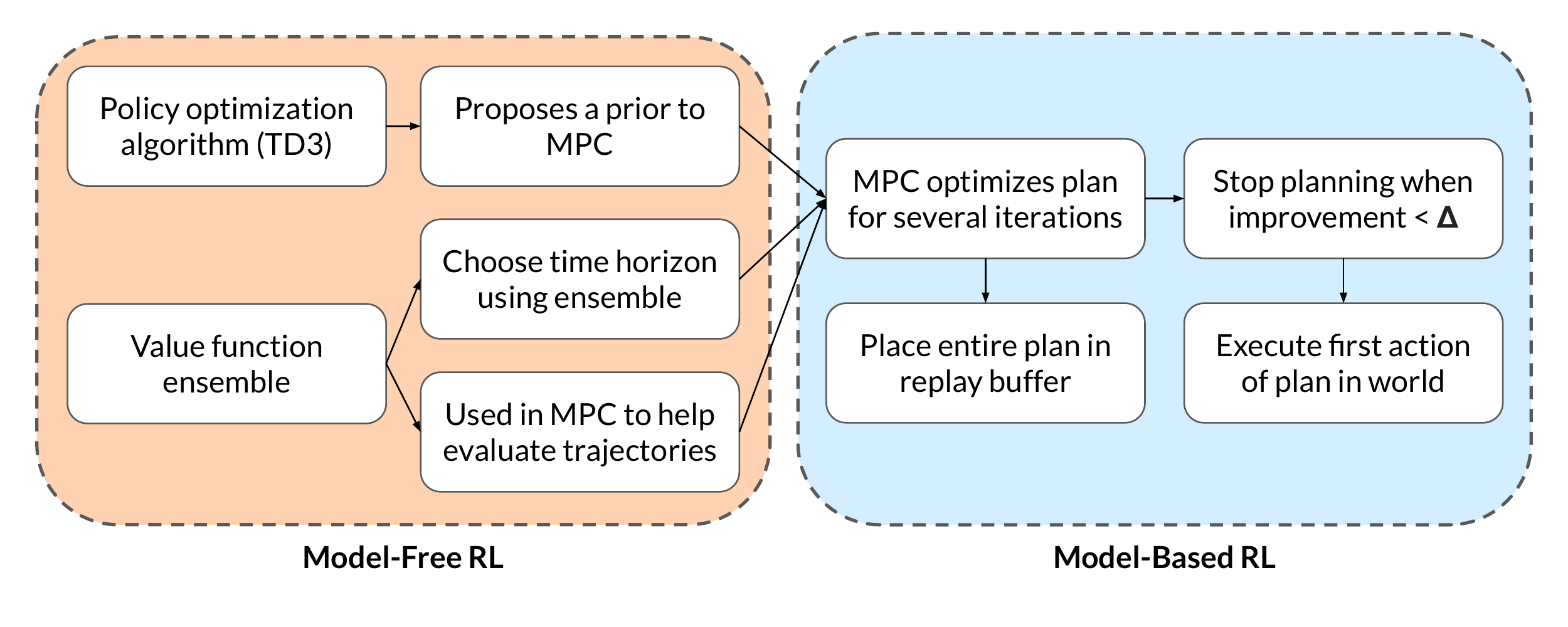}
    \caption{Schematic view of Adaptive Online Planning (AOP).}
    \label{fig:aopdiag}
\end{figure}

\begin{algorithm}[H]
    \caption{Adaptive Online Planning (AOP)} \label{alg:aoplong}
    Initialize neural networks: value ensemble $\{V_i\}_{i=1}^n$, policy prior $\pi_\theta$ \\
    Initialize value replay buffer $\mathcal{D}_V$, policy replay buffer $\mathcal{D}_\pi$ \\
    \While{alive} {
    
        \texttt{\\}
        \tcc{Off-Policy Prior (Section \ref{sec:priors})}
        Generate $\tau_\pi = (s_t, \pi_\theta(s_t), s_{t+1}, \pi_\theta(s_{t+1}), ..., s_{t+H})$ from policy $\pi_\theta$ \\
        Set $\tau_{plan} := \argmax_{\tau_\pi, \tau_{plan}} \hat{J}^\pi_\infty(\tau)$ \\
        
        \texttt{\\}
        \tcc{Adaptive Planning Horizon (Section \ref{sec:thor})}
        Calculate the standard deviation $\sigma$ of $\{V_i(s_t)\}_{i=1}^n$ and Bellman error $\eps(H | \tau_{plan})$ \\
        \If{$\sigma < \sigma_{thres}$} {
            Set $H_t$ equal to the maximum $H < H_{full}$ such that all $\eps(H | \tau_{plan}) > \eps_{thres}$
        }
        \Else {
            Set $H_t := H_{full}$
        }
        
        \texttt{\\}
        \tcc{Model-Based Planning with Value Approximation (Section \ref{sec:polo})}
        Set $\tau_1 := \tau_{plan}$ \\
        \For{$k\gets1$ \KwTo $max\_iters$} {
            Generate $pop\_size$ action sequences $\{(a^i_t, a^i_{t+1}, ..., a^i_{t+H_t-1})\}_i$ by adding noise to $\tau_{k}$ \\
            Roll out each sequence in the model to calculate the value $\hat{J}^\pi_\infty(s_t)$ of each trajectory \\
            Add rolled out sequences to replay buffer $\mathcal{D}_\pi$ \\
            Calculate $\tau_{k+1}$ using MPPI softmax weighting with each weight $w_i \propto e^{\frac{1}{\lambda} \hat{J}^\pi_\infty(\tau_i)}$ \\
            
            \texttt{\\}
            \tcc{Early Planning Termination (Section \ref{sec:ept})}
            Calculate plan improvement $\Delta(\tau_{k+1} | \tau_k)$ \\
            \If{$\Delta(\tau_{k+1} | \tau_k) < \Delta_{thres}$} {
                Terminate planning with probability $1 - \eps_{plan}$
            }
        }
        
        \texttt{\\}
        \tcc{Model-Free Learning}
        \If{it's time to update} {
            Update value ensemble $\{V_i\}_{i=1}^n$ with $\mathcal{D}_V$ \\
            Update policy $\pi_\theta$ with $\mathcal{D}_\pi$ using policy optimization algorithm
        }
        Step once in the environment with the first action of the new trajectory $\tau_{final}$ \\
        Add $(s_t, a_t, s'_t, r_t)$ to replay buffer $\mathcal{D}_V$ \\
        Set $\tau_{plan}$ to be equal to $\tau_{final}$ advanced by one timestep \\
    }
\end{algorithm}

\section{Hyperparameters}

Our hyperparameters are chosen coarsely from similar work, and new hyperparameters were approximately chosen based on performance in the 2D maze environment. We found most of the hyperparameters to not be very sensitive, except for the MPPI temperature, which is important for all algorithms and not specific to AOP.

\subsection{Adaptive Online Planning Hyperparameters} \label{app:aophyperparameters}

For AOP, we set $\sigma_{thres} = 8$, $\eps_{thres} = 25$ and $\eps_{plan} = 0.2$. For the first planning iteration we set $\Delta_{thres} = 0.01$, and for the later planning iterations, $\Delta_{thres} = 0.05$, which helps to prioritize the use of the planner. These values were coarsely chosen by looking at general values of the value function, and we did not tune these hyperparameters much. We found that the algorithm is not overly sensitive to the thresholds in dense reward environments, which we also show in Appendix \ref{app:sensitivity}. The only exception to this is that we set $\sigma_{thres} = 1$ for the Humanoid environment, as the value function is harder to learn accurately. In the sparse reward settings, we set $\sigma_{thres} = \eps_{thres} = 0$, in order to avoid early termination of exploration (we do not change the hyperparameters determining the number of planning iterations). This is a natural concession that is hard to resolve in the reset-free setting as the exploration in the problem is not as well-defined, and must be addressed by some broader change to the setting, which we leave for future work.

\subsubsection{Sensitivity to Thresholds} \label{app:sensitivity}

\begin{figure}[h]
    \centering
    \includegraphics[scale=0.4]{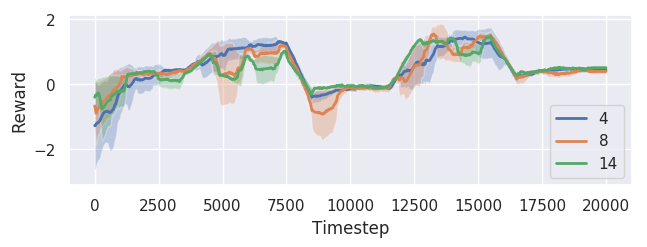} \includegraphics[scale=0.4]{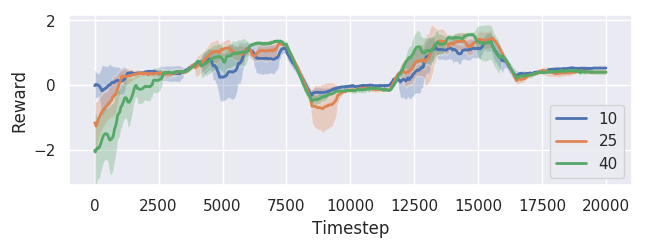}
    \caption{Learning curves for hyperparameter sweep. Left: standard deviation $\sigma_{thres}$. Right: Bellman error $\epsilon_{thres}$. Legend shows value of relevant hyperparameter. 13 seeds were run in total.}
    \label{fig:hyp-robust}
\end{figure}

To test the sensitivity of AOP to the heuristic thresholds from Section \ref{sec:thor}, we run a rough grid search with wider values for $\sigma_{thres}$ and $\eps_{thres}$ in the Hopper changing worlds environment. The average reward for each setting is shown in Table \ref{table:hyp-robust} and learning curves are shown in Figure \ref{fig:hyp-robust}. AOP is somewhat more sensitive to the setting of $\eps_{thres}$ early on in training, as a higher value corresponds to less planning, but this effect quickly dissipates. As a result, while the choice of $\sigma_{thres}$ and $\eps_{thres}$ is somewhat arbitrary, we do not believe that AOP is particularly sensitive to them.

\begin{table}[h]
\caption{Effect of varying threshold hyperparameters}
\label{table:hyp-robust}
\begin{center}
\begin{tabular}{llllllll}
\multicolumn{1}{c}{\bf Standard Deviation $\sigma_{thres}$:}  & \multicolumn{1}{c}{\bf 4} & \multicolumn{1}{c}{\bf 8 (Default)} & \multicolumn{1}{c}{\bf 14} 
\\ \hline \\
Average reward & 0.47 $\pm$ 0.20 & 0.42 $\pm$ 0.05 & 0.44 $\pm$ 0.16 \\ \\
\multicolumn{1}{c}{\bf Bellman Error $\epsilon_{thres}$:}  & \multicolumn{1}{c}{\bf 10} & \multicolumn{1}{c}{\bf 25 (Default)} & \multicolumn{1}{c}{\bf 40}
\\ \hline \\
Average reward & 0.47 $\pm$ 0.17 & 0.47 $\pm$ 0.09 & 0.43 $\pm$ 0.24
\end{tabular}
\end{center}
\end{table}

\subsubsection{Uncertainty Heuristics in the WhyNot Environment}

\begin{figure}[hbt!]
    \centering
    \includegraphics[scale=0.2]{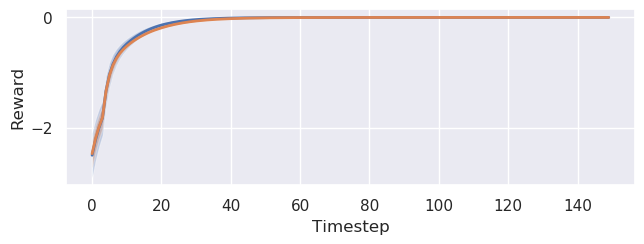}
    \includegraphics[scale=0.2]{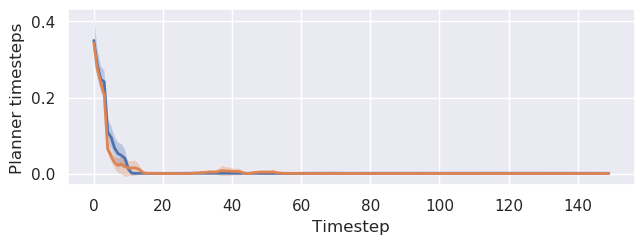}
    \\
    \includegraphics[scale=0.2]{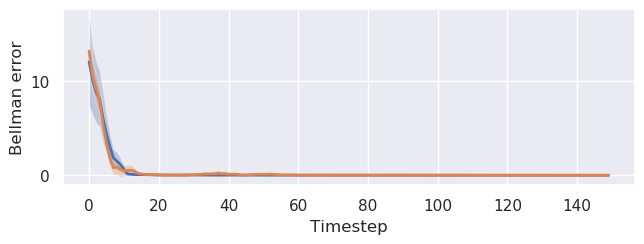}
    \includegraphics[scale=0.2]{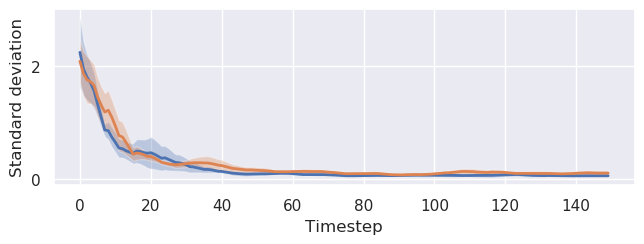}
    \caption{AOP results on the WhyNot Zika environment. Top: performance metrics (reward, planning timesteps). Bottom: uncertainty metrics (Bellman error, ensemble standard deviation).}
    \label{fig:zikametrics}
\end{figure}

To consider whether the heuristics generalize outside of our physics-based settings, we also test AOP in the causal WhyNot simulator \cite{miller20whynot} on its continuous control Zika task. Using the same hyperparameters, AOP achieves optimal performance ($\approx$ MPC), shown in Table \ref{table:zikaperf}. The agent correctly plans less once converged, reflected in the uncertainty metrics, shown in Figure \ref{fig:zikametrics}.

\begin{table}[h]
\caption{Lifetime Performance on WhyNot Zika}
\label{table:zika}
\begin{center}
\begin{tabular}{llll}
\multicolumn{1}{c}{\bf Algorithm:} & \multicolumn{1}{c}{\bf AOP-TD3} & \multicolumn{1}{c}{\bf AOP-BC} & \multicolumn{1}{c}{\bf MPC-3}
\\ \hline \\
Average reward & -0.12 $\pm$ 0.05 & -0.13 $\pm$ 0.02 & -0.14 $\pm$ 0.01
\end{tabular}
\end{center}
\label{table:zikaperf}
\end{table}

\subsection{Model Predictive Control Hyperparameters}

Our MPPI temperature $\l$ is set to 0.01. The other planning hyperparameters are shown below. See Section \ref{sec:priors} for interpretation of policy optimization as a special case of AOP. Surprisingly, we found TD3 to perform worse with more than 1 trajectory per iteration.

\begin{table}[h]
\label{table:plan-hyp}
\begin{center}
\begin{tabular}{llllll}
\multicolumn{1}{c}{\bf Parameter}  & \multicolumn{1}{c}{\bf AOP-TD3/AOP-BC} & \multicolumn{1}{c}{\bf POLO} & \multicolumn{1}{c}{\bf TD3} & \multicolumn{1}{c}{\bf PPO} & \multicolumn{1}{c}{\bf MPC}
\\ \hline \\
Planning horizon & 1-80 & 80 & 256 & 128 & 80 \\
Planning iterations per timestep & 0-8 & 3 & 1 & 1 & 3, 8 \\
Trajectories per iteration & 40 & 40 & 1 & 32 & 40 \\
Noise standard deviation & 0.1 & 0.1 & 0.2 & - & 0.1 \\
\end{tabular}
\end{center}
\end{table}

For the Humanoid environment only, all the MPC-based algorithms use 120 trajectories per iteration and a noise standard deviation of $1$.

\subsection{Network Architectures}

For our value ensembles, we use an ensemble size of $6$ and $\kappa = 10^{-2}$. The value functions are updated in batches of size $32$ for $32$ gradient steps every $4$ timesteps. All networks use tanh activations and a learning rate of $10^{-3}$, trained using Adam. Network sizes are shown below.

\begin{table}[!htbp]
\label{table:networksizes}
\begin{center}
\begin{tabular}{lllll}
\multicolumn{1}{c}{\bf Environment} & \multicolumn{1}{c}{\bf AOP-TD3/AOP-BC} & \multicolumn{1}{c}{\bf POLO} & \multicolumn{1}{c}{\bf TD3} & \multicolumn{1}{c}{\bf PPO}
\\ \hline \\
Hopper & $V: (64,64), Q: (400,300), \pi: (400, 300)$ & $(64,64)$ & $(400,300)$ & $(64,64)$ \\
Humanoid & $V: (64,64), Q: (400,300), \pi: (400, 300)$ & $(64,64)$ & $(400,300)$ & $(64,64)$ \\
Ant & $V: (64,64), Q: (400,300), \pi: (400, 300)$ & $(64,64)$ & $(400,300)$ & $(64,64)$ \\
Maze & $(64,64)$ & $(64,64)$ & $(64,64)$ & $(64,64)$ \\
\end{tabular}
\end{center}
\end{table}

\subsection{Policy Optimization Hyperparameters}

Our TD3 uses the same hyperparameters as the original authors \cite{td3}, where for every timestep, we run a rollout of length 256 and run 256 gradient steps. In the TD3 used for the episodic policy learning experiment in Section \ref{sec:polbehavior}, we run rollouts of length 1000 and run 1000 gradient steps after each rollout, equivalent to the standard TD3 setting with no terminal states.

Our PPO uses $\eps=0.2$, $\l = 0.95$, batch sizes of $4096$, and $80$ gradient steps per iteration, which we found to yield the best results out of a set of common hyperparameters. For behavior cloning, we run $400$ gradient steps on batches of size $64$ every $4$ timesteps. For the policy in AOP-TD3, we run $128$ gradient steps on batches of size $100$ every $4$ timesteps.

\section{Environment Details} \label{app:envs}

\begin{figure*}[h]
    \centering
    \includegraphics[scale=0.35]{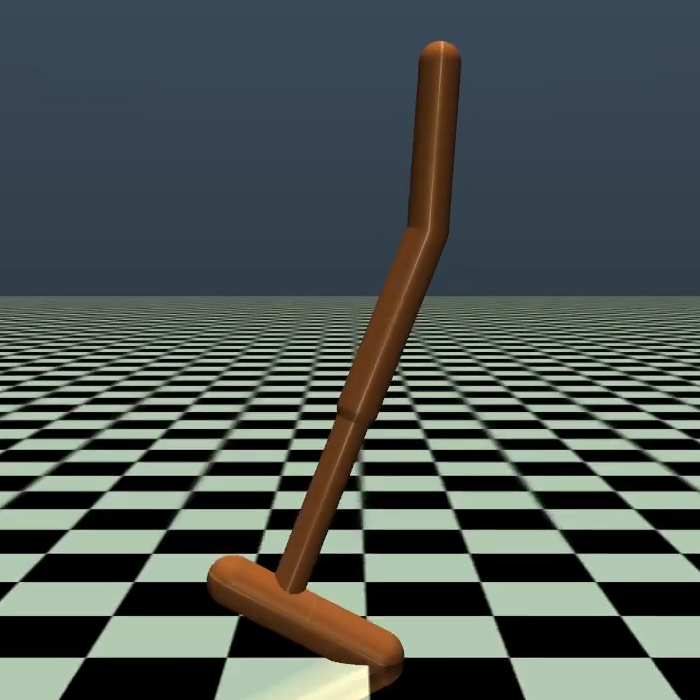} \qquad
    \includegraphics[scale=0.147]{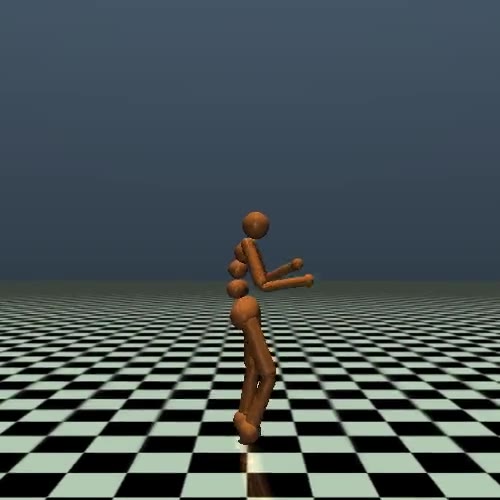} \qquad
    \includegraphics[scale=0.35]{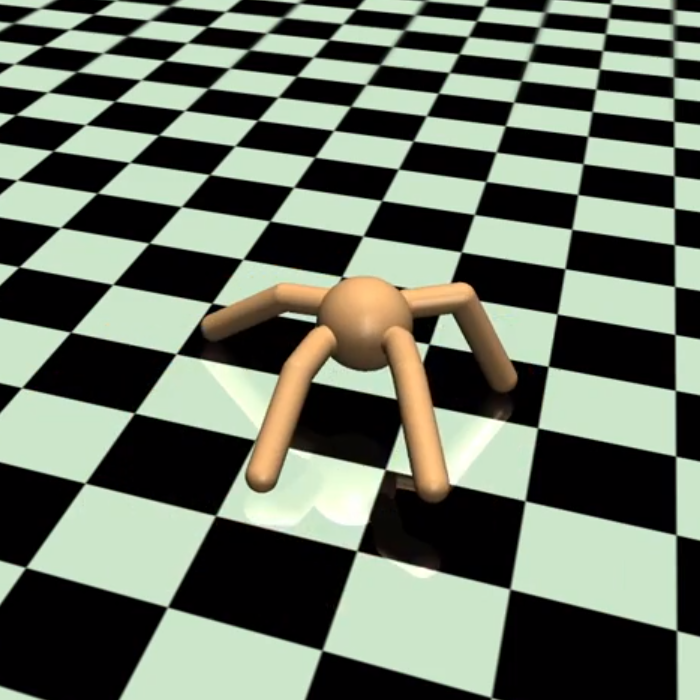} \qquad
    \includegraphics[scale=0.7]{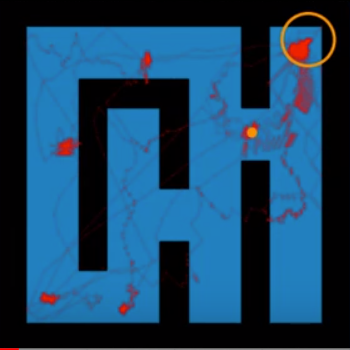}
    \caption{Pictures of lifelong learning environments (from left to right): Hopper, Humanoid, Ant, and Maze. In the Maze, the agent (orange ball) must try to navigate to the goal (open orange circle) while avoiding the black walls. In the picture, the red lines indicate past history of movement.}
    \label{fig:envs}
\end{figure*}

\subsection{Environment Descriptions}

\textbf{Hopper}: We use the Hopper agent from OpenAI Gym \cite{gym}. The agent is tasked with achieving a target forward $x$-velocity (in the range of $1-3$) in the following settings:
\begin{itemize}
    \item Standard (S): the target velocity is constant
    \item Novel states (NS): the target velocity changes, but is observed by the state
    \item Changing worlds (CW): the target velocity changes and is not observed by the state
\end{itemize}

\textbf{Humanoid}: We use the SlimHumanoid agent from OpenAI Gym \cite{gym}, which reduces the dimension of the state space significantly. The agent is tasked with achieving a target forward $x$-velocity while the mass of its right arm varies (in the range of $50-300\%$):
\begin{itemize}
    \item Standard (S): the head mass is constant
    \item Changing worlds (CW): the head mass varies and is not observed by the state
\end{itemize}

\textbf{Ant}: We use the Ant agent from OpenAI Gym \cite{gym}, with the full state space. The agent is tasked with achieving a target forward $x$-velocity while a joint at random is disabled (e.g. the action is set to $0$):
\begin{itemize}
    \item Standard (S): no joints are disabled
    \item Changing worlds (CW): the disabled joint varies and is not observed by the state
\end{itemize}

\textbf{Maze}: We use a 2D particle maze environment. The walls of the maze change and the goal swaps between two locations to avoid the agent staying still at the goal forever. The goal is always observed, making the observation space 4D. We test on both dense and sparse reward variants.
\begin{itemize}
    \item Novel states (NS): the walls are constant, but the new unseen goals appear periodically
    \item Changing worlds (CW): the goals stay in the same two places, but the walls change
\end{itemize}

\subsection{Reward Specifications}

In the online setting, the agent receives no signal from termination states, i.e. it becomes more difficult to know not to fall down in the cases of Hopper and Ant. This leads to a ``rolling'' behavior for the Hopper and a ``wiggling'' behavior for the Ant, which, while these modes can capable of achieving the same velocities, are not particularly meaningful or difficult to achieve. To amend this, and achieve the same stable/interpretable behavior as the standard reinforcement learning setting, we set the reward functions as the following for our environments, in the style of \cite{polo, simple}:

\begin{table}[h]
\begin{center}
\begin{tabular}{lllllll}
\multicolumn{1}{c}{\bf Environment} & \multicolumn{1}{c}{\bf Reward Function}
\\ \hline \\
Hopper & $|x\_vel - x\_vel_{targ}| + 5 (z-1.8)^2 + .1 \N{a}_2^2 + x\_vel_{targ}$ \\
Humanoid & $1 - \min(1.1, z) - 0.25 * |x\_vel - 1| + ctrl\_cost$ \\
Ant & $|x\_vel - 2| + 3 (z-.9)^2 + .01 \N{a}_2^2$ \\
Maze (Dense) & $-\N{(x,y) - (x,y)_{goal}}_2 - \mathbf{1}\{\text{contact with wall}\}$ \\
Maze (Sparse) & $\mathbf{1}\{\text{inside goal}\} - \mathbf{1}\{\text{contact with wall}\}$ \\
\end{tabular}
\end{center}
\end{table}

\section{Detailed Experimental Graphs} \label{app:graphs}

\begin{figure}[h]
    \centering
    \includegraphics[scale=1.2]{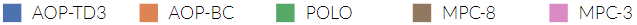} \\
    \begin{minipage}{.32\textwidth}
        \centering
        \includegraphics[scale=0.27]{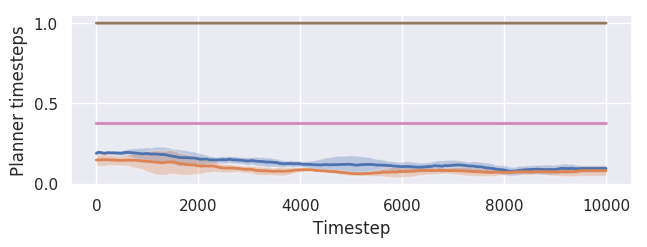} \\
        (a) Standard Hopper
    \end{minipage}
     \begin{minipage}{.32\textwidth}
        \centering
        \includegraphics[scale=0.27]{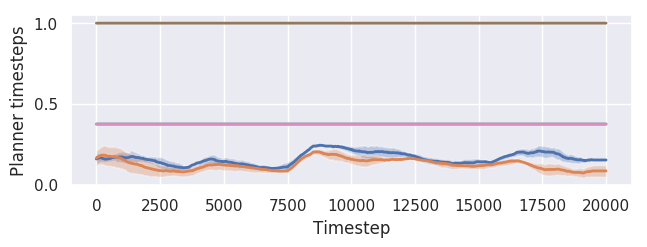} \\
        (b) Novel states Hopper
    \end{minipage}
     \begin{minipage}{.32\textwidth}
        \centering
        \includegraphics[scale=0.27]{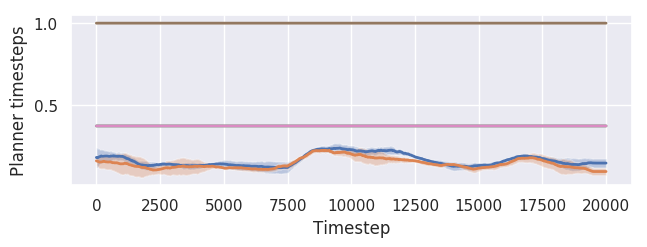} \\
        (c) Changing worlds Hopper
    \end{minipage}
    \begin{minipage}{.32\textwidth}
        \centering
        \includegraphics[scale=0.27]{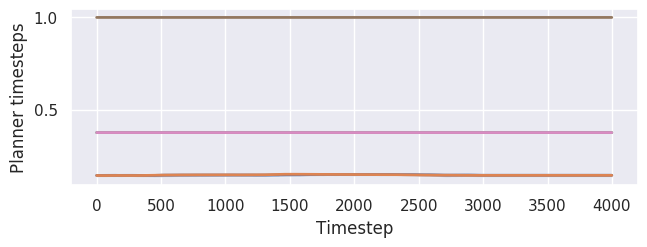} \\
        (d) Standard Humanoid
    \end{minipage}
     \begin{minipage}{.32\textwidth}
        \centering
        \includegraphics[scale=0.27]{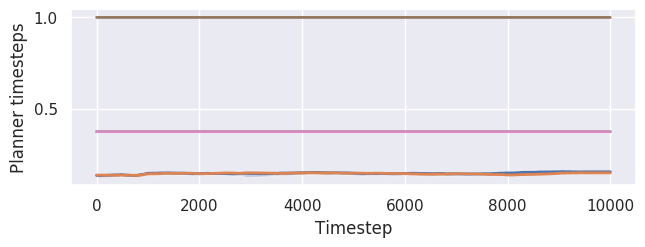} \\
        (e) Changing worlds Humanoid
    \end{minipage}
    \begin{minipage}{.32\textwidth}
        \centering
        \includegraphics[scale=0.27]{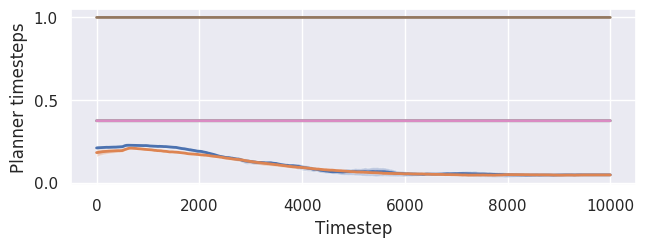} \\
        (f) Standard Ant
    \end{minipage}
    \begin{minipage}{.32\textwidth}
        \centering
        \includegraphics[scale=0.27]{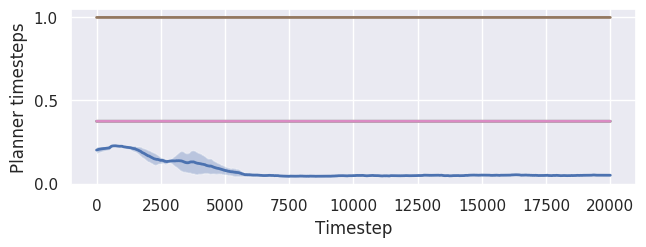} \\
        (g) Changing worlds Ant
    \end{minipage}
    \begin{minipage}{.32\textwidth}
        \centering
        \includegraphics[scale=0.27]{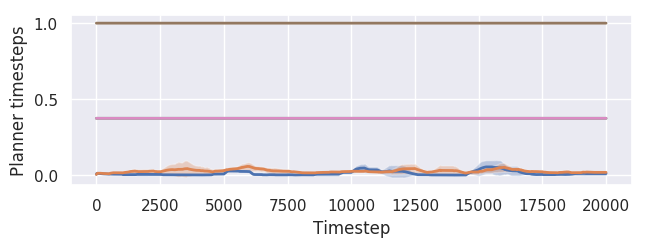} \\
        (h) Novel states Maze (D)
    \end{minipage}
    \begin{minipage}{.32\textwidth}
        \centering 
        \includegraphics[scale=0.27]{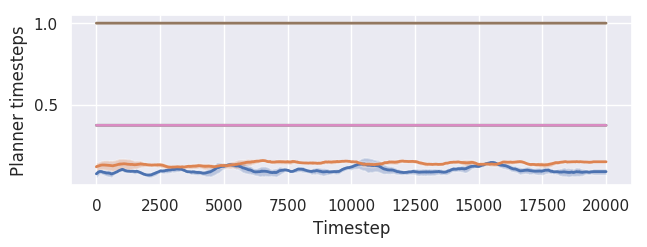} \\
        (i) Changing worlds Maze (D)
    \end{minipage}
     \begin{minipage}{.32\textwidth}
        \centering
        \includegraphics[scale=0.27]{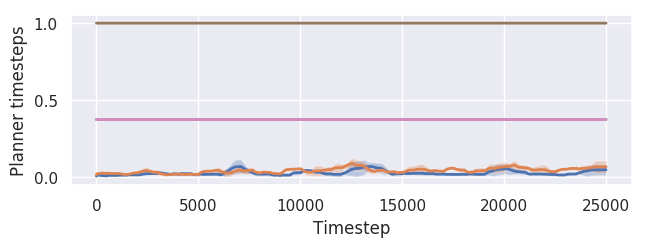} \\
        (j) Novel states Maze (S)
    \end{minipage}
     \begin{minipage}{.32\textwidth}
        \centering
        \includegraphics[scale=0.27]{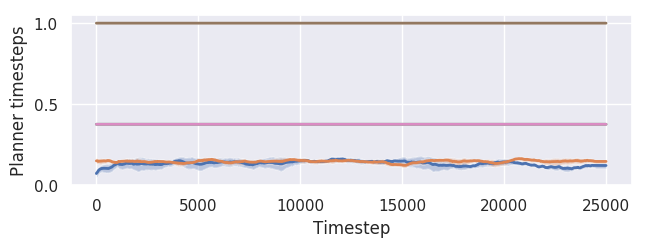} \\
        (k) Changing worlds Maze (S)
    \end{minipage}
    \caption{Number of timesteps rolled out by planner per timestep as a percentage of MPC-8.}
    \label{fig:psteps}
\end{figure}

\begin{figure}[h]
    \centering
    \includegraphics[scale=1.2]{pics/algo_labels}
    \begin{minipage}{.32\textwidth}
        \centering
        \includegraphics[scale=0.27]{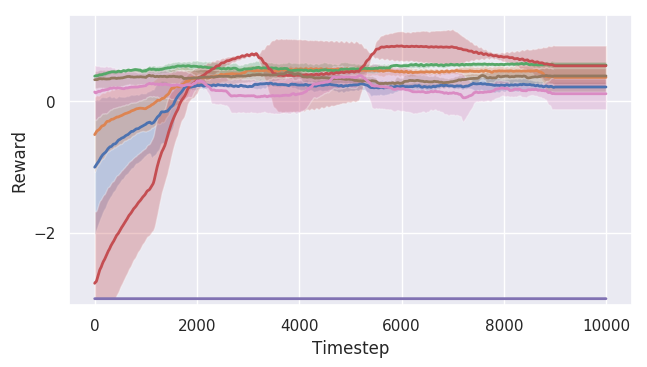} \\
        (a) Standard Hopper
    \end{minipage}
     \begin{minipage}{.32\textwidth}
        \centering
        \includegraphics[scale=0.27]{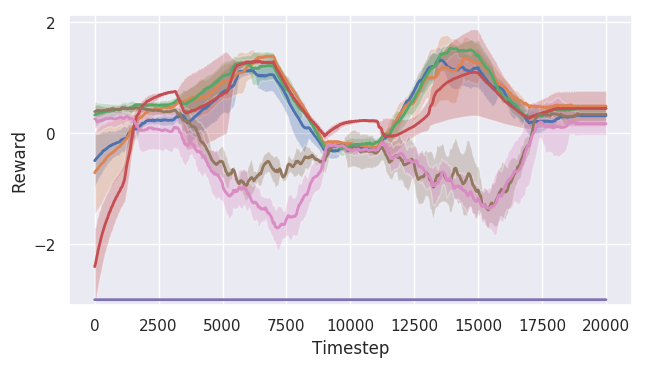} \\
        (b) Novel states Hopper
    \end{minipage}
     \begin{minipage}{.32\textwidth}
        \centering
        \includegraphics[scale=0.27]{pics/rewards/hop-c.png} \\
        (c) Changing worlds Hopper
    \end{minipage}
    \begin{minipage}{.32\textwidth}
        \centering
        \includegraphics[scale=0.27]{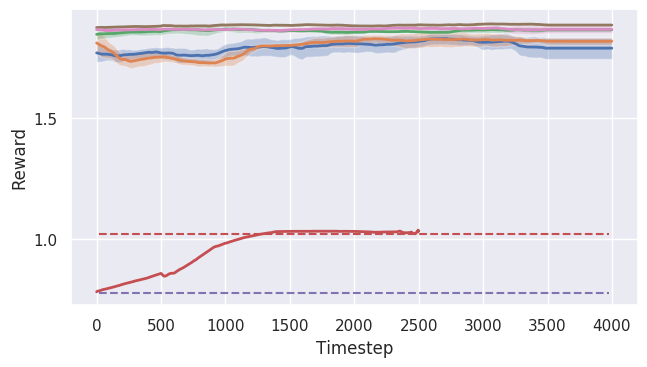} \\
        (d) Standard Humanoid
    \end{minipage}
     \begin{minipage}{.32\textwidth}
        \centering
        \includegraphics[scale=0.27]{pics/rewards/humanoid-c.png} \\
        (e) Changing worlds Humanoid
    \end{minipage}
    \begin{minipage}{.32\textwidth}
        \centering
        \includegraphics[scale=0.27]{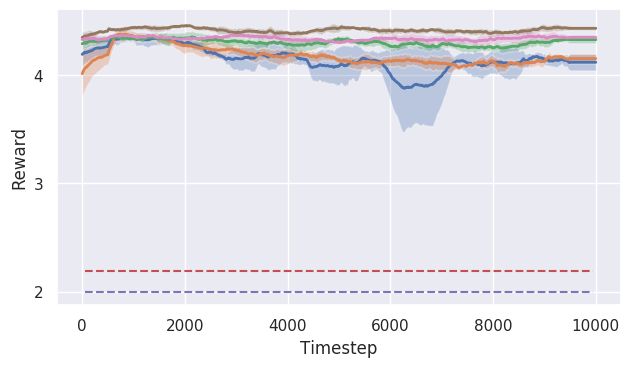} \\
        (f) Standard Ant
    \end{minipage}
    \begin{minipage}{.32\textwidth}
        \centering
        \includegraphics[scale=0.27]{pics/rewards/ant-c.png} \\
        (g) Changing worlds Ant
    \end{minipage}
    \begin{minipage}{.32\textwidth}
        \centering
        \includegraphics[scale=0.27]{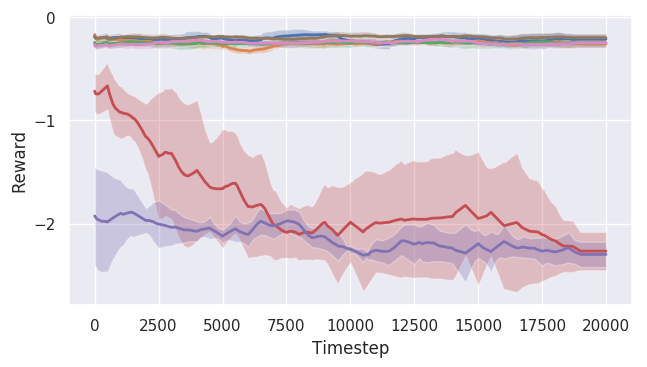} \\
        (h) Novel states Maze (D)
    \end{minipage}
    \begin{minipage}{.32\textwidth}
        \centering 
        \includegraphics[scale=0.27]{pics/rewards/maze-c.png} \\
        (i) Changing worlds Maze (D)
    \end{minipage}
    \begin{minipage}{.32\textwidth}
        \centering
        \includegraphics[scale=0.27]{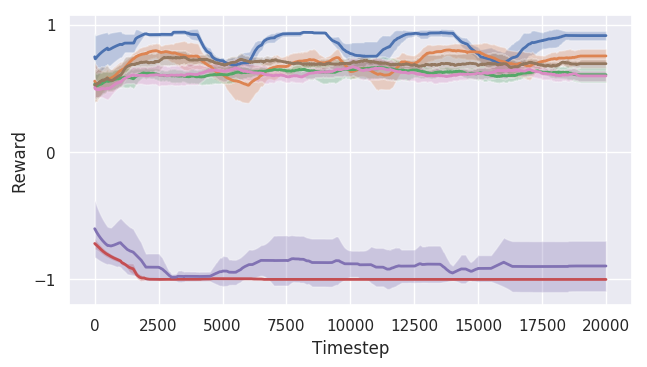} \\
        (j) Novel states Maze (S)
    \end{minipage}
     \begin{minipage}{.32\textwidth}
        \centering
        \includegraphics[scale=0.27]{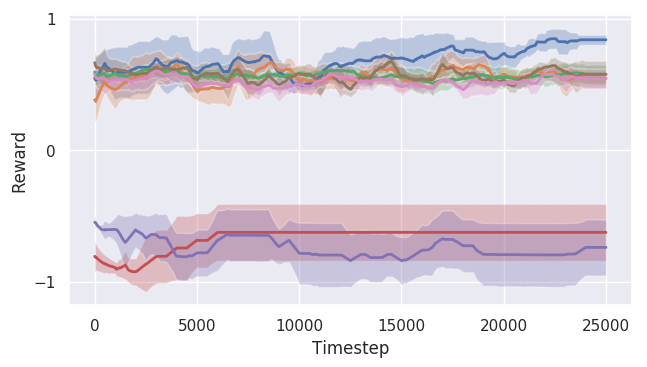} \\
        (k) Changing worlds Maze (S)
    \end{minipage}
    
    \caption{Reward curves for lifelong learning tasks.}
    \label{fig:rew}
\end{figure}

Additional graphs are provided for the fraction of planning timesteps vs MPC-8 and reward curves for all environments in Figures \ref{fig:psteps} and \ref{fig:rew}, respectively.


\end{document}